\documentclass{WileyMSP-template}

\usepackage{upgreek}
\usepackage{graphicx}
\usepackage{epstopdf}
\usepackage{dcolumn}
\usepackage{bm}
\usepackage{multirow}
\usepackage[cmyk]{xcolor}
\usepackage{soul}
\usepackage{upgreek}
\usepackage{longtable}
\usepackage{amsmath}
\usepackage{amssymb}
\usepackage{cleveref}
\crefname{figure}{\textcolor{blue!80!black}{Figure}}{\textcolor{blue!80!black}{Figure}}
\crefname{equation}{\textcolor{blue!80!black}{Equation}}{\textcolor{blue!80!black}{Equation}}
\crefname{table}{\textcolor{blue!80!black}{Table}}{\textcolor{blue!80!black}{Table}}

\usepackage[misc]{ifsym}
\usepackage[T1]{fontenc}

\usepackage{booktabs}
\usepackage{multirow}
\usepackage{diagbox}
\usepackage[version=3]{mhchem}
\usepackage{enumerate}
\usepackage{sidecap}
\usepackage{booktabs}
\usepackage{tabularx}

\usepackage{lineno}

\newcommand*\E{\mathbb{E}}

\begin{document}


\pagestyle{fancy}

\title{Utilizing Causal Network Markers to Identify Tipping Points ahead of Critical Transition}

\maketitle


\author{Shirui Bian}
\author{Zezhou Wang}
\author{Siyang Leng}
\author{Wei Lin}
\author{Jifan Shi*}

\begin{affiliations}
S. Bian, Prof. W. Lin\\
School of Mathematical Sciences\\
Fudan University\\
Shanghai 200433, China

Z. Wang, Prof. W. Lin\\
Shanghai Center for Mathematical Sciences\\
Fudan University\\
Shanghai 200433, China

S. Bian, Prof. S. Leng, Prof. W. Lin, Prof. J. Shi\\
Research Institute of Intelligent Complex Systems\\
Fudan University\\
Shanghai 200433, China\\
Email: jfshi@fudan.edu.cn

Prof. S. Leng\\
Institute of AI and Robotics, Academy for Engineering and Technology\\
Fudan University\\
Shanghai 200433, China\\

Prof. W. Lin\\
Shanghai Artificial Intelligence Laboratory\\
Shanghai 200232, China

Prof. W. Lin\\
State Key Laboratory of Medical Neurobiology and MOE Frontiers Center for Brain Science\\
Fudan University\\
Shanghai 200032, China

\end{affiliations}



\begin{abstract}

  Early-warning signals of delicate design are always used to predict critical transitions in complex systems, which makes it possible to render the systems far away from the catastrophic state by introducing timely interventions. Traditional signals including the dynamical network biomarker (DNB), based on statistical properties such as variance and autocorrelation of nodal dynamics, overlook directional interactions and thus have limitations in capturing underlying mechanisms and simultaneously sustaining robustness against noise perturbations. This paper therefore introduces a framework of causal network markers (CNMs) by incorporating causality indicators, which reflect the directional influence between variables. Actually, to detect and identify the tipping points ahead of critical transition, two markers are designed: CNM-GC for linear causality and CNM-TE for non-linear causality, as well as a functional representation of different causality indicators and a clustering technique to verify the system's dominant group. Through demonstrations using benchmark models and real-world datasets of epileptic seizure, the framework of CNMs shows higher predictive power and accuracy than the traditional DNB indicator. It is believed that, due to the versatility and scalability, the CNMs are suitable for comprehensively evaluating the systems. The most possible direction for application includes the identification of tipping points in clinical disease.

\end{abstract}


\section{Introduction}

When numerous similar entities at micro-level engage in interactions among themselves and with their environment, spontaneous and often unexpected outcomes emerge at spatiotemporal macro-level. This phenomenon, known as emergence, is a hallmark of various complex systems \cite{Artime2022PT}. Emergence is frequently linked to collective behavior, wherein many living systems demonstrate critical transition \cite{Scheffer2001Nature,Scheffer2009Nature,Lenton2008PNAS}. At these tipping points, systems undergo transitions from normal states to catastrophic states. As a result, extensive investigations have been dedicated to designing early-warning signal and thus predicting the occurrence of such critical transitions between different states within these systems \cite{Wissel1984,Chen2012SR}. Early-warning signals of delicate design are essential for predicting critical transitions in complex dynamical systems, which makes it possible to conduct timely interventions, prevent catastrophic events and mitigate negative impacts. Some traditional methods, e.g., the dynamical network biomarker (DNB), rely on statistical quantities such as variance, autocorrelation, and dimension reduction, which offers valuable insights into the system's state \cite{Scheffer2009Nature,Chen2012SR,Liu2014Bio,Liu2015SR,Jiang2018PNAS}. However, such methods have difficulties in capturing the directional interactions and underlying mechanisms that drive the dynamics of complex systems.

Causality indicates a directional influence between a pair of variables and is a fundamental form of interactions within complex systems. Researchers have developed various sophisticated causality indicators to characterize the relationships between nodes in complex networks, such as the Granger causality (GC), the transfer entropy (TE), and the embedding entropy (EE) \cite{Granger1969,Shi2022Interface,Ying2022Research}. These indicators, due to their broad applicability and diversity, hold significant potential for enhancing our understanding of system dynamics in different senses. For instance, approaches based on Taken's embedding theorem effectively characterize the relationship between non-linearly coupled nodes and solve a critical causal inference problem in terms of non-separability \cite{Shi2022Interface,Sugihara2012Science}. Inspired by these advances, incorporating causality indicators into the framework of early-warning signals may provide a more robust and informative approach for identifying critical points and predicting system bifurcations.

In this paper, we proposed a framework of causal network markers (CNMs) to detect typical critical bifurcations in the dynamical evolution of complex systems. Our framework begins with the $K$-means clustering method based on data variance, categorizing nodes into dominant group (DG) and non-dominant group (NDG). More importantly, our theoretical analysis shows that as the system nears a tipping point, the unidirectional GC from DG nodes to NDG nodes vanishes. Leveraging this property, we construct CNMs to indicate the extent to which the dynamics approaches a tipping point. Specifically, we select GC and TE under the framework of CNM, viz., CNM-GC, which quantifies ``linear causality'', and CNM-TE, which quantifies ``non-linear causality''. 

To validate the efficacy of CNM-GC and CNM-TE, we applied them to both benchmark models and real-world datasets. The benchmark models include a five-gene network, an ecological network, and a Turing diffusion interaction network, each representing different critical phenomena. Our results demonstrate that CNMs have significant predictive power for both temporal and spatial bifurcation models. In addition, we tested our markers using the epileptic seizure dataset, intracranial electroencephalography (iEEG), where the combination of the two markers showed high predictive accuracy. It is worth noting that due to the complexity of epilepsy dynamics, combining CNM-GC and CNM-TE information could further identify the types of causal state that trigger the tipping point. By integrating multiple causal indicators, CNMs offer a comprehensive identification of system dynamics, making them suitable for applications in clinical disease detection and early-warning.

\section{Results and Discussions}

\subsection{Vanishment of Granger causality near the tipping point}

Here, we illustrate that when the system's dynamics approaches a tipping point, the causality from the DG nodes to the NDG nodes vanishes, in the sense of Granger causality. Generally, a discrete dynamical system with multidimensional parameter $\boldsymbol P$ is governed by
\begin{equation}\label{}
	\begin{aligned}
		\boldsymbol Z^{t+1} = \boldsymbol{f}(\boldsymbol Z^t; \boldsymbol P),
	\end{aligned}
\end{equation}
where $\boldsymbol Z^t=(z_1^t,z_2^t,\dots,z_n^t)^\top$ is a vector with $n$ components, and $\boldsymbol{f}$ is a $C^1$-function with a non-trivial fixed point $\bar{\boldsymbol Z}$, i.e., $\boldsymbol{f}(\bar{\boldsymbol Z})=\bar{\boldsymbol Z}$. Assume there exists a tipping point at parameter $\boldsymbol P_c$, where the system undergoes a codimension-one bifurcation. Linearization of the system around $\bar{\boldsymbol Z}$ yields
\begin{equation}\label{}
	\begin{aligned}
		\boldsymbol X^{t+1}=\boldsymbol A\boldsymbol X^t+\boldsymbol \Gamma^t,
	\end{aligned}
\end{equation}
where $\boldsymbol X^t=\boldsymbol Z^t-\bar{\boldsymbol Z}$,
\begin{equation}\label{}
	\begin{aligned}
		\boldsymbol A=\left.\frac{\partial f(\boldsymbol Z; \boldsymbol P)}{\partial \boldsymbol Z}\right\lvert_{\boldsymbol Z=\bar{\boldsymbol Z}}
	\end{aligned}
\end{equation}
is the Jacobian matrix of $\boldsymbol{f}$ at $\bar{\boldsymbol Z}$, and $\boldsymbol \Gamma^t=(\Gamma_1^t,\dots,\Gamma_n^t)^\top$ contains all remaining high-order terms. To directly display our conclusion, we write down a typical situation where $\boldsymbol A$ can be diagonalized in $\mathbb{R}^{n\times n}$: $\boldsymbol A=\boldsymbol S\boldsymbol \Lambda \boldsymbol S^{-1}$ with $\boldsymbol S,\,\boldsymbol \Lambda\in\mathbb{R}^{n\times n}$. Set $\boldsymbol Y^t=\boldsymbol S^{-1}\boldsymbol X^t$ and we can obtain
\begin{equation}\label{}
	\begin{aligned}
		&\boldsymbol S\boldsymbol Y^{t+1}=\boldsymbol A\boldsymbol S\boldsymbol Y^t+\boldsymbol \Gamma^t=(\boldsymbol S\boldsymbol \Lambda \boldsymbol S^{-1})\boldsymbol S\boldsymbol Y^t+\boldsymbol \Gamma^t,\\
		& \boldsymbol Y^{t+1}=\boldsymbol \Lambda \boldsymbol Y^t+\boldsymbol S^{-1}\boldsymbol \Gamma^t,
	\end{aligned}
\end{equation}
where $\boldsymbol Y^t=(y_1^t,y_2^t,\dots,y_n^t)^\top$ satisfies a diagonal dynamic. In the following discussion, we assume $y_1$ to be the only one ``dominant variable'' in this diagonal dynamic. That is, when $\boldsymbol P$ goes to $\boldsymbol P_c$, the variance of $y_1$ tends to infinity, while the Pearson correlation coefficients (PCC) between $y_1$ and other variables tend to 0 \cite{Chen2012SR}. Now we calculate the GC between any two variables $x_i$ and $x_j$ in the original phase space \cite{Granger1969}.

For $x_i^t=s_{i1}y_1^t+\dots+s_{in}y_n^t$ and $x_j^t=s_{j1}y_1^t+\dots+s_{jn}y_n^t$, there are $H_0$-model and $H_1$-model for the causality from $x_j$ to $x_i$:
\begin{equation}\label{model}
	\begin{cases}
		\begin{aligned}
			H_0:\,\, x_i^{t+1}=ax_i^t+\xi_1,
		\end{aligned}\\
		\begin{aligned}
			H_1:\,\, x_i^{t+1}=bx_i^t+cx_j^t+\xi_2.
		\end{aligned}
	\end{cases}
\end{equation}
Based on \cref{model}, we prove that when $s_{i1}=0$ and $s_{j1}\neq 0$, i.e., the $j$-th variable is related to the dominant variable $y_1^t$ while the $i$-th not, $\E[\xi_2^2]\rightarrow \E[\xi_1^2]$ holds as the variance of $y_1$ tends to infinity. Thus, the Granger causality strength from $x_j^t$ to $x_i^t$, $\text{GC}_{x_j\rightarrow x_i}=\log \frac{\E[\xi_1^2 ]}{\E[\xi_2^2]}$, tends to 0. Besides, we discuss the other three situations (see detailed proof in Supporting Information). The conclusions for the four cases are summarized as
\begin{enumerate}[(a)]
	\item $s_{i1}=0$, $s_{j1}\neq 0$, and $\text{GC}_{x_j\rightarrow x_i}$ tends to 0;

	\item $s_{i1}\neq0$, $s_{j1}= 0$, and $\text{GC}_{x_j\rightarrow x_i}$ is bounded;

	\item $s_{i1}\neq 0$, $s_{j1}\neq 0$, and $\text{GC}_{x_j\rightarrow x_i}$ is bounded;

	\item $s_{i1}=0$, $s_{j1}= 0$, and $\text{GC}_{x_j\rightarrow x_i}$ is invariant.

\end{enumerate}

The results indicate that when the system attains a tipping point, the causality from the variable corresponding to the dominant diagonalized variables (i.e., the variables in DG), to those in NDG tends to be 0, while the causality from the variables in NDG to those in DG changes within a bounded number. Meanwhile, the causality between variables in NDG does not change.

\begin{SCfigure*}
	\centering
	\includegraphics[width=0.8\textwidth]{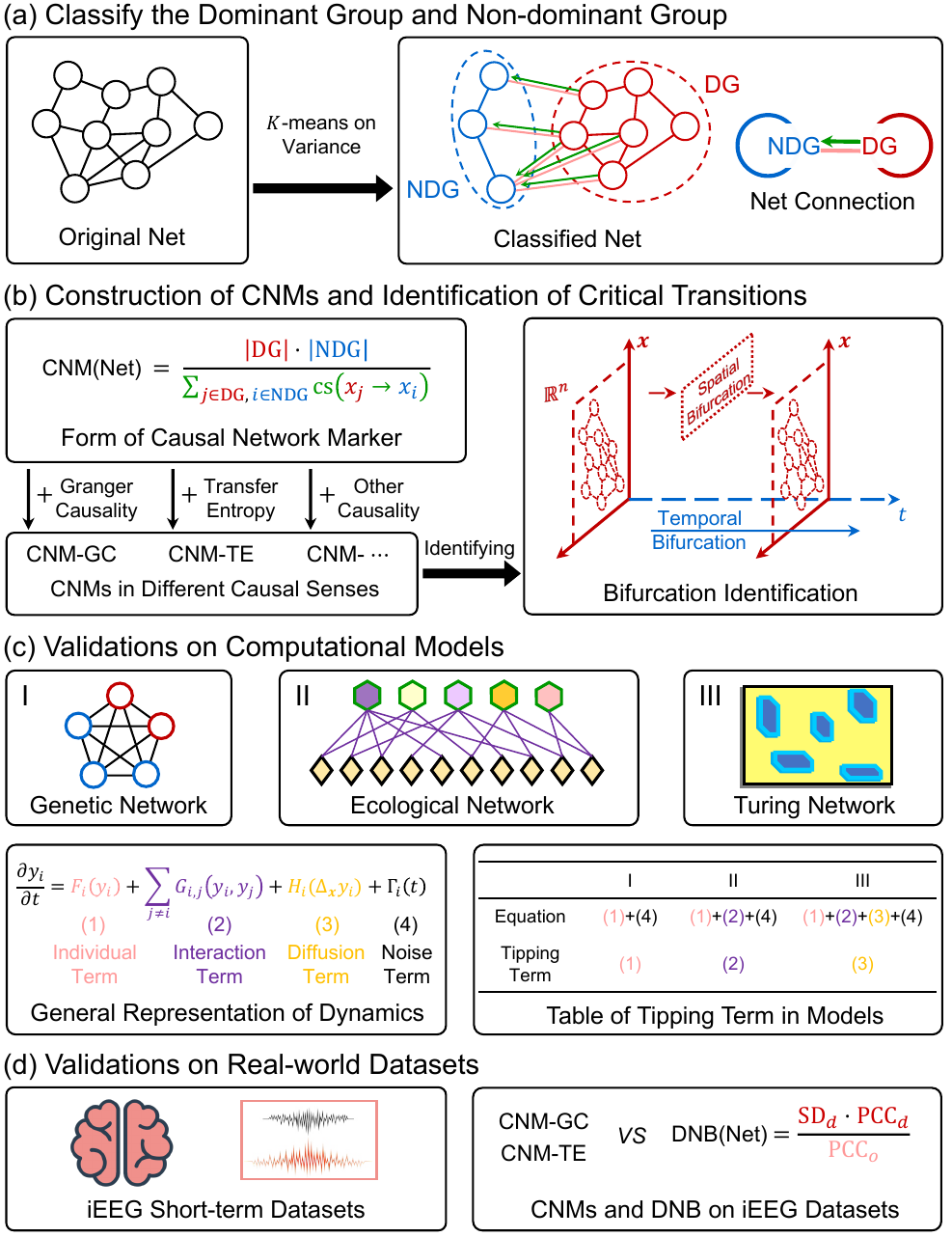}
	\caption{A sketch depicting the main procedure for calculating the CNMs. (a) Verifying the DG and the NDG through the $K$-means algorithm. (b) Calculating the CNMs with different causality methods, such as GC and TE. (c) CNMs' validation on detecting the general bifurcation phenomena on benchmark models. (d) CNMs' validation on the iEEG seizure dataset, exhibits a more scalable and flexible approach to detecting the tipping point of real-world datasets.}
	\label{Figure_1}
\end{SCfigure*}

\subsection{A framework of causal network markers for tipping point identification}

In the previous section, it is shown that the causality strength of GC rapidly changes when the system evolves in the vicinity of the tipping point. Empirically, GC reflects causality in a ``linear sense''. Naturally, we can extend the idea to a broader sense of causality, such as TE and other causal indicators.

TE is a non-parametric statistic that measures the amount of directional (time asymmetric) information transfer between two random processes. For Gaussian processes, TE is simplified as GC. In the general sense, the TE is designed to reveal causality in the ``non-linear sense''. Specifically, the calculation of TE is given in the following form: $\text{TE}_{X\rightarrow Y} = H\left( Y_{t+1} \mid Y_t\right) - H\left( Y_{t+1} \mid Y_t, X_t\right)$, where $H(\cdot|\cdot)$ represents the conditional entropy. Therefore, in this work, we not only use GC, but also introduce TE.

Based on the previous proof of the ``causal vanishing'' property, we propose a network marker framework with GC or TE to detect critical transitions, which can be easily extended to other types of causality (such as EE), in principle. Firstly, we divide nodes into DG and NDG via the $K$-means algorithm, where the clusters are obtained by the average variance of nodes during the given period (see pesudocode in Supporting Information). After that, our marker is calculated based on the clusters as follows:
\begin{equation}
    \text{CNM(Net)} := \frac{\lvert \text{DG} \lvert \cdot \lvert \text{NDG} \lvert}{\sum_{j \in \text{DG}, i \in \text{NDG}}\text{cs}\left( x_j \rightarrow x_i \right)},
\end{equation}
where $\lvert \cdot \lvert$ represents the number of the elements in a set, and $\text{cs}(\cdot)$ is the causal strength in the sense of GC, TE, or other types of causality.

Through the CNM, if there is an element approaching 0 in the set of $\{x_j\in \text{DG},\,x_i\in\text{NDG} \, \lvert \, \text{cs}(x_j\rightarrow x_i)\rightarrow0\}$, the marker blows up. Therefore, we use the significant increase in the marker as an effective early-warning signal for tipping points. In \cref{Figure_1}, we sketch the major procedure for the CNM framework.

\subsection{Validations on benchmark models}

Here, we apply the CNM framework combined with two types of causality to benchmark models. From the complex biological networks' perspective, the bifurcation of system evolution usually has different internal causes, led by different terms in the dynamical equations \cite{Ashwin2012PT,Schoenmakers2021Chaos}. Specifically, the representation of complex dynamics with $n$ nodes is governed as \cite{Barabasi2016nature}:
\begin{equation}\label{}
	\begin{aligned}
		\frac{\partial y_i}{\partial t}=&F_i(y_i)+\sum_{j\neq i}G_{i,j}(y_i,y_j)\\
		&+H_i(\Delta_{\boldsymbol{x}} y_i)+\Gamma_i(t),\,i=1,2,\dots,n,
	\end{aligned}
\end{equation}
where $y_i=y_i(t;\boldsymbol{x})$ is a temporal-spatial variable, $F_i$ is the individual evolving dynamics, $G_{i,j}$ is the interaction term between nodes, $H_i$ is the spatial diffusion term for the spatial states $\boldsymbol{x}$, and $\Gamma_i$ is Gaussian white noise with:
\begin{equation}\label{}
	\begin{aligned}
		\left\langle \Gamma_i(t) \right\rangle=0,\,\left\langle \Gamma_i(t)\Gamma_j(t') \right\rangle= 2D\delta_{ij}\delta(t-t').
	\end{aligned}
\end{equation}

To test the effectiveness of our framework under different types of bifurcation phenomena, three typical biological networks are selected from the application, namely, the genetic networks, the ecological networks, and the Turing networks. Numerical experiments show that CNM-GC and CNM-TE can identify these simple network models with different types of bifurcation phenomena effectively.

\subsubsection{Five-gene genetic network}

The five-gene genetic network is a regulatory network and the system is governed by 5-dimensional Langevin equations \cite{Chen2002IEEE, Yuh1998Sci, Chen2012SR}:
\begin{equation}\label{}
	\begin{cases}
		\begin{aligned}
			\frac{\text{d}z_1(t)}{\text{d}t} =& (90\lvert P\lvert -1236)+\frac{240-120\lvert P\lvert}{1+z_3(t)}\\
			&+\frac{1488z_4(t)}{1+z_4(t)}-30\lvert P\lvert z_1(t)+\Gamma_1(t),
		\end{aligned}\\
		\begin{aligned}
			\frac{\text{d}z_2(t)}{\text{d}t} =& (75\lvert P\lvert-150)+\frac{60-30\lvert P\lvert}{4z_1(t)-2}\\
			&+\frac{(240-120\lvert P\lvert)z_3(t)}{1+z_3(t)}-60z_2(t)+\Gamma_2(t),
		\end{aligned}\\
		\begin{aligned}
			\frac{\text{d}z_3(t)}{\text{d}t} = -1056+\frac{1488z_4(t)}{1+z_4(t)}-60z_3(t)+\Gamma_3(t),
		\end{aligned}\\
		\begin{aligned}
			\frac{\text{d}z_4(t)}{\text{d}t} = -600+\frac{1350z_5(t)}{1+z_5(t)}-100z_4(t)+\Gamma_4(t),
		\end{aligned}\\
		\begin{aligned}
			\frac{\text{d}z_5(t)}{\text{d}t} =& 108+\frac{160}{1+z_1(t)}+\frac{40}{1+z_2(t)}\\
			&+\frac{1488}{1+z_4(t)}-300z_5(t)+\Gamma_5(t).
		\end{aligned}
	\end{cases}
\end{equation}

\begin{figure*}[t!]
	\centering
	\includegraphics[width=0.91\textwidth]{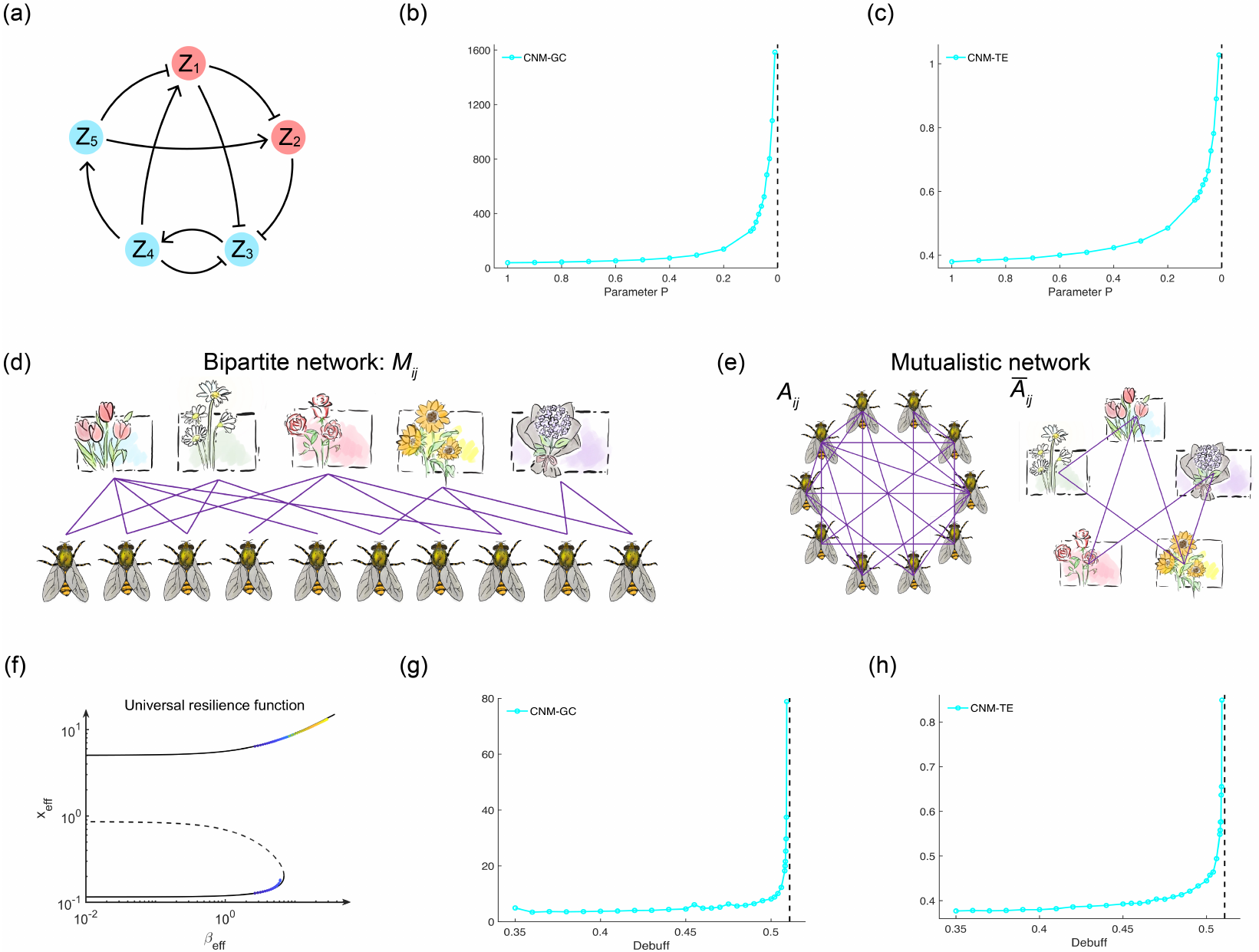}
	\caption{Early-warning signals of the five-gene genetic network and the mutualistic interaction network based on the causality marker. (a) The network connection of the genetic network, with the theoretically verified DG (red nodes): $\{Z_1,\,Z_2\}$. (d,e) The network connection of the mutualistic interaction network, where the separate networks (e) of the pollinators and the plants are generated from the bipartite interaction network (d). (b,g) In the sense of GC, the CNM increases rapidly when $P$ approaches 0. (c,h) In the sense of TE, the conclusion is similar. (f) Universal resilience function obtained from our numerical simulation, which shows that Barabasi's dimension reduction of dynamics is reasonable.}
	\label{Figure_2}
\end{figure*}

In this genetic circuit, the eigenvalues of its corresponding discrete linearized system at the unique stable steady state $\bar{\boldsymbol Z}=(1,0,1,3,2)$ are
\begin{equation}\label{}
	\begin{aligned}
		(0.74^{\lvert P \lvert},0.55,0.37,0.20,0.14),
	\end{aligned}
\end{equation}
which indicates that a phase transition occurs when $P\rightarrow 0$. Additionally, we identify ${z_1,z_2}$ as the DG corresponding to the largest eigenvalue $\lambda_{\text{max}}=0.74^{\lvert P \lvert}$, which governs the phase transition ($\lambda_{\text{max}}\rightarrow 1$ as $\lvert P \lvert\rightarrow 0$) as described in \cite{Chen2012SR}. In theory, the causality strength will change and can be denoted by CNM. To empirically investigate this property, we conduct numerical simulations. The results reveal that both CNM values exhibit pronounced trends, offering early-warning signals for the impending phase transition (see \cref{Figure_2}(a-c)).

\subsubsection{Ecological mutualistic interaction network}

In addition to bifurcations resulting from individual parameter changes, it is also essential to consider the occurrence of bifurcation when the interaction undergoes variations. A typical mutualistic interaction network comprises two distinct species such as pollinators, denoted as $PO_{j}$ with $j=1,\dots,n$, and plants, denoted as $PL_{i}$ with $i=1,\dots,m$. Their interactions are characterized by bipartite relationships represented by the matrix $(M_{ij})_{\substack{1\leq i\leq n\\1\leq j\leq m}}$ (\cref{Figure_2}(d)). After that, we can construct two distinct mutualistic networks by establishing connections between pairs of plants and pollinators derived from the $M_{ij}$ matrix, following the methodology outlined in \cite{Barabasi2016nature}:
\begin{equation}\label{}
	\begin{aligned}
		&A_{ij}=\frac{\sum\limits_{k=1}^m M_{ik}M_{jk}}{\sum\limits_{s=1}^n M_{sk}},\, 1\leq i,j\leq n,\\
		&\bar{A}_{ij}=\frac{\sum\limits_{k=1}^n M_{ki}M_{kj}}{\sum\limits_{s=1}^m M_{km}},\,1\leq i,j\leq m.
	\end{aligned}
\end{equation}
Here, $A_{ij}\, (1\leq i,j\leq n)$ denotes the interaction strength within the pollinators network, while $\bar{A}_{ij}\,(1\leq i,j\leq m)$ denotes the interaction strength within the plants network (\cref{Figure_2}(e)). Notably, the dynamical influence of one species on another is implicitly encapsulated within the parameters $A_{ij}$ or $\bar{A}_{ij}$. Consequently, we can construct the dynamics of both the plants' network and the pollinators' network as follows (we only display the pollinators' network) \cite{Barabasi2016nature,Gao2020interface,Gao2022NE}:
\begin{equation}\label{dynamic_barabasi}
	\begin{aligned}
		\frac{\text{d}x_i}{\text{d}t}=&s\left[B_i+x_i\left(1-\frac{x_i}{K_i}\right)\left(\frac{x_i}{C_i}-1\right)\right.\\
		&+\left.\sum\limits_{j=1}^n A_{ij}\frac{x_ix_j}{D_i+E_ix_i+H_jx_j}\right]+\Gamma_i(t),\\
		&i=1,\dots,n.
	\end{aligned}
\end{equation}
In this dynamic, the first term accounts for logistic growth, encompassing factors such as the Allee effect and a constant influx attributed to migration. These factors are characterized by parameters $B_i$, $K_i$, $C_i$, $D_i$, $E_i$, $H_i$, and a scaling parameter $s$ (where $s=1$ in the work by Barabasi). The interaction term quantifies the symbiotic influence of each population $x_j$ on $x_i$, with its strength determined by the parameter $A_{ij}$. This interaction term saturates as the population sizes become sufficiently large.

Detailed investigations of such networks have been conducted by Gao et al. \cite{Barabasi2016nature, Gao2020interface, Gao2022NE}. Their research findings have revealed that when the parameters in \cref{dynamic_barabasi} exhibit node-independence, with values such as $B_i=B$, $C_i=C$, $D_i=D$, $E_i=E$, $H_i=H$, and $K_i=K$, the high-dimensional dynamics can be effectively reduced to one-dimensional resilient dynamics through the application of mean-field approximation:
\begin{equation}\label{ODE_eff}
	\begin{aligned}
		\frac{\text{d}x_{\text{eff}}}{\text{d}t}=&B+x_{\text{eff}}\left(1-\frac{x_{\text{eff}}}{K}\right)\left(\frac{x_{\text{eff}}}{C}-1\right)\\
		&+\beta_{\text{eff}}\frac{x_{\text{eff}}^2}{D+(E+H)x_{\text{eff}}}.
	\end{aligned}
\end{equation}
In this context, we denote $x_{\text{eff}}=\frac{\boldsymbol{1}^\top \boldsymbol{A}\boldsymbol{x}}{\boldsymbol{1}^\top \boldsymbol{A}\boldsymbol{1}}$ as the efficient state, while $\beta_{\text{eff}}$ represents a critical parameter referred to as the universal resilient state. Actually, $\beta_{\text{eff}}$ serves to aggregate the influence of the interaction matrix $\boldsymbol{A}$. Specifically, within the framework of mean-field approximation, we can express $\beta_{\text{eff}}$ as $\beta_{\text{eff}}=\frac{\boldsymbol{1}^\top\boldsymbol{A}^2 \boldsymbol{1}}{\boldsymbol{1}^\top\boldsymbol{A} \boldsymbol{1}}$ \cite{Barabasi2016nature}. Subsequently, we can solve the ODE \cref{ODE_eff} that governs the dynamics of $x_{\text{eff}}$ and $\beta_{\text{eff}}$. This equation yields $\beta_{\text{eff}}=\beta_{\text{eff}}(x_{\text{eff}})$, which is regarded as a resilience function, describing the appropriate behavior of the stochastic evolution (\cref{Figure_2}(f)). Barabasi and colleagues have observed a bifurcation phenomenon occurring when the interaction strengths are perturbed globally as $A_{ij}\rightarrow w_{ij}A_{ij}$. In this scenario, when both the efficient state $x_{\text{eff}}$ and the universal resilient state $\beta_{\text{eff}}$ simultaneously reach a threshold denoted as $x_{\text{eff}}^c$ and $\beta_{\text{eff}}^c$, respectively, defined as:
\begin{equation}\label{thresholds}
\begin{aligned}
\left.\frac{\partial f(\beta_{\text{eff}}^c,x_{\text{eff}})}{\partial x_{\text{eff}}}\right\lvert_{x_{\text{eff}}^c}=0,\
f(\beta_{\text{eff}}^c,x_{\text{eff}}^c)=0,
\end{aligned}
\end{equation}
the system undergoes a transition from either a bistable or monostable state to the other. Specifically, the stability of one steady state, indicative of ``low population density'', changes to either manifest or vanish. 

In our numerical simulations, we have considered a specific scenario where the pollinators are considered invasive alien species, necessitating strict population control measures to maintain a small population size. Consequently, it becomes crucial to generate an early-warning signal before the low population steady state loses its stability. To adapt our framework effectively to this context, we have intentionally disregarded the stochastic of $w_{ij}$ and introduced a weak Gaussian white noise component into the dynamics to incorporate stochastic effects. In our initial parameter selection, as defined in \textcolor{blue!80!black}{Tab.} S1, we aimed to establish stability in the high state while rendering the low state unstable. Subsequently, we deliberately set $w_{ij}=\text{Debuff}\equiv 0.3$ to restore stability to both states. We systematically increased this global weight adjustment to examine the parameters that lead to instability in the low state. It's worth noting that under the influence of the noise term, the threshold is not a fixed constant but exhibits fluctuations within a small range around $\text{Debuff}=0.52$. This observation implies that as the low state becomes increasingly unstable, invasive alien species may proliferate, posing a significant threat to the ecological system.

Our analysis of the time series data obtained from simulations allows us to calculate the CNM under both GC and TE measures, revealing a consistent surge in CNM values preceding the bifurcation (see \cref{Figure_2}(d-h)). Furthermore, we calculate the efficient state $x_{\text{eff}}$ and its corresponding universal resilient state $\beta_{\text{eff}}$ using simulations, as outlined in \cref{ODE_eff} (\cref{Figure_2}(f)). In instances where the system exhibits monostability for smaller values of $\text{Debuff}$ or bistability for larger values of $\text{Debuff}$, our approximations of $x_{\text{eff}}$ and $\beta_{\text{eff}}$ consistently cluster near the resilience function. This observation underscores the rationality and effectiveness of Barabasi's dimension reduction approach to capturing the system's dynamics \cite{Barabasi2016nature}. In summary, our results demonstrate the versatility of the CNM in detecting early-warning signals for systems undergoing bifurcations, whether they are induced by single parameter changes, or global parameter changes, as exemplified by ecological mutualistic interaction networks. Notably, early-warning signals in such multifactorial systems pose a unique challenge due to the unfixed nature of their tipping points, influenced by a multitude of parameters. These complex, multifactorial dynamics may not be effectively detected by conventional early-warning signals. However, the CNM provides a precise and comprehensive signal, as it encompasses a range of intricate and implicit influencing factors inherent in causality indices. This makes it well-suited for detecting bifurcations driven by multifactorial influences.

\subsubsection{Turing diffusion interaction network}
\begin{figure*}[t!]
	\centering
	\includegraphics[width=0.95\textwidth]{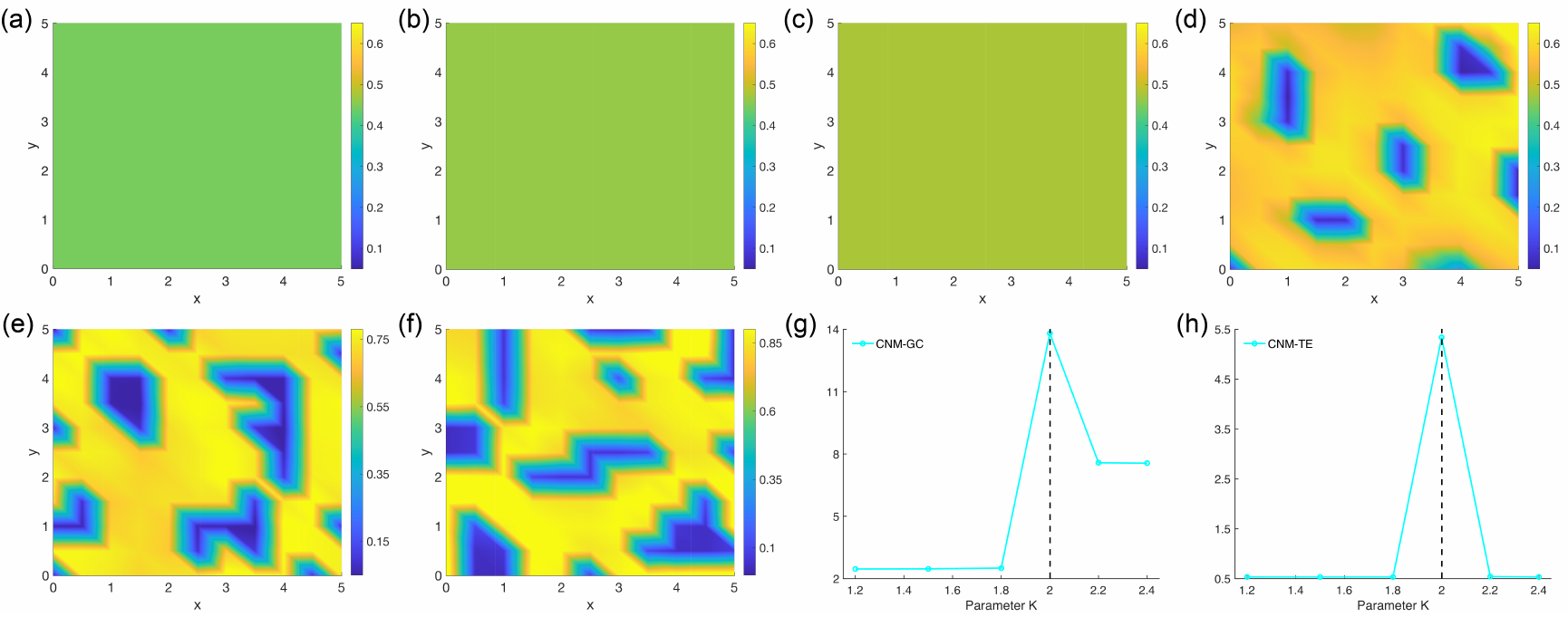}
	\caption{Early-warning signals of the Turing diffusion interaction network based on the causality marker. (a-f) The spatial broken phenomenon occurs near the Turing bifurcation. The transition parameter $K$ is set as (a) $K$=1.2, (b) $K$=1.5, (c) $K$=1.8, (d, the pattern occurs) $K$=2, (e) $K$=2.2, and (f) $K$=2.4. (g-h) In the sense of GC and TE, the CNMs increase rapidly only when the bifurcation occurs ($K\approx 2$).}
	\label{Figure_3}
\end{figure*}

Lastly, we explore the efficacy of the CNM in detecting spatial bifurcation, highlighting its versatile applications. Beyond morphogenesis theories in developmental biology, reaction-diffusion systems with spatial heterogeneity are relevant in various contexts. A reaction-diffusion system shows diffusion-driven instability, or Turing instability, when the homogeneous steady state remains stable without diffusion but becomes unstable to spatial perturbations with diffusion present.

In biology, instability often refers to the situation where a uniform steady state destabilizes under small perturbations, leading to non-uniform behavior of ecological significance. Examples include environmental heterogeneity in animal dispersal \cite{Pickett1995Science}, reaction-diffusion in anisotropic growth domains \cite{Krause2019BMB}, spatial invasion modelling \cite{Belmonte2013JTB}, and differential diffusion in plant root initiation \cite{Avitabile2018SIAM}. Spatial heterogeneity influences local instability conditions, modulates pattern size and wavelength, and localizes spike patterns. Here, we focus on the Turing bifurcations, a complex phenomenon where the system transitions from a homogeneous stable state to a non-uniform stable state under specific conditions.

The predator-prey model, which represents many realistic biological phenomena, also displays spatially non-uniform behavior during the Turing instability. It is proposed (In this work, we call it Turing network for convenience) by Beddington and DeAngelis in a form of \cite{Beddington1975JAE}:
\begin{equation}\label{}
	\begin{cases}
		\begin{aligned}
			\frac{\partial H(t,x,y)}{\partial t}=&r\left(1-\frac{H}{K}\right)H-\frac{\beta H}{B+H+\omega P}P\\
			&+D_1\Delta_{x,y} H+\Gamma_1(t),
		\end{aligned}\\
		\begin{aligned}
			\frac{\partial P(t,x,y)}{\partial t}=&\frac{\varepsilon\beta H}{B+H+\omega P}P-\eta P\\
			&+D_2\Delta_{x,y} P+\Gamma_2(t),
		\end{aligned}
	\end{cases}
\end{equation}
where the Turing bifurcation occurs with significantly different diffusion coefficients $D_1$ and $D_2$. The parameters are set as $r=0.5$, $\varepsilon=1$, $\beta=0.6$, $B=0.4$, $\eta=0.25$, $\omega=0.4$, and $D_2=1$ with $\Delta_{x,y} = \partial^2_x + \partial^2_y$. The remaining parameters $D_1$ and $K$ control the spatial pattern.

The spatial pattern arises from the instability of the time-invariant steady state $H^*(x,y)$ and $P^*(x,y)$ that satisfy:
\begin{equation}\label{}
	\begin{cases}
		\begin{aligned}
			r\left(1-\frac{H^*}{K}\right)H^*-\frac{\beta H^*}{B+H^*+\omega P^*}P^*+D_1\Delta_{x,y} H^*=0,
		\end{aligned}\\
		\begin{aligned}
			\frac{\varepsilon\beta H^*}{B+H^*+\omega P^*}P^*-\eta P^*+D_2\Delta_{x,y} P^*=0.
		\end{aligned}
	\end{cases}
\end{equation}
when $D_1$ and $D_2$ differ significantly, the spatially homogeneous structure breaks, producing patterns with different expression levels at different spatial locations. In numerical simulations, we divide the spatial range $(x,y) \in [0,R_x] \times [0,R_y]$ into $11\times 11$ lattices, and treating each lattice as a variable. We then apply our detection procedure to this system in the sense of CNM-GC and CNM-TE.

In the study of the Turing networks, minor differences in parameter selection leads to significant variations in the system's dynamical behavior, particularly the emergence of the Turing-Hopf bifurcations \cite{Tang2016ND,Chen2023EN}. Relevant researches about this system have illustrated the bifurcation conditions in the parameter space, with regions representing the different types of bifurcation behaviors the system may experience \cite{Tang2016ND}. Specifically, the area below the Hopf bifurcation line indicates that the system remains stable over time without oscillations; however, the area above suggests the onset of periodic oscillatory behavior. Similarly, the area below the Turing bifurcation line implies spatial uniformity in the system, while the area above indicates the formation of spatial patterns. After the Turing bifurcation, with parameter changes, the system may exhibit a variety of pattern formations. Near the bifurcation point, the system generally displays a spot pattern structure, while far from the bifurcation point, it typically shows a stripe pattern \cite{Dakos2011AN,Kefi2014PLOS,Krause2020Interface}.

Through the analysis of genetic networks and ecological networks, we find that bifurcations in the temporal dimension can be effectively detected by our CNM-GC and CNM-TE indices. However, the core of this subsection is to demonstrate that these indices can also detect spatial bifurcation phenomena such as Turing bifurcations. To this end, we select specific parameter combinations $D_1=0.01,\,D_2=1$. As the parameter $D_1$ increases, the system gradually enters Turing bifurcation without accompanying Hopf bifurcation.

In terms of experimental design, we conduct simulation experiments for each set of parameters $(D_1, D_2)$ for up to 800 seconds to ensure that the system's evolution is in the vicinity of equilibrium. We employ a common data aggregation strategy: Integrating continuous simulation data with a 1-second time window, thereby treating the node distribution at each second as a snapshot of the data within that second. The rationale for this method is based on the assumption that over a sufficiently short time scale, the system's dynamical behavior exhibits a certain consistency, allowing us to approximate the node distribution characteristics throughout the simulation process through short-time data aggregation.

Furthermore, unlike the previous two systems where the DG was known, in this system with 121 lattices, we cannot a priori know the DG. Therefore, we use the $K$-means clustering algorithm to distinguish between DG and NDG. \cref{Figure_3} shows the trend of changes in the CNM-GC and CNM-TE indices under different parameter settings. It can be observed from the figure that both markers reach their peaks near the middle, close to the parameters where Turing bifurcation occurs, which validates the effectiveness of the markers, that is, the system's tipping dynamics promote an increase in the CNM.

It is particularly noteworthy that the CNM-GC index shows a certain increase after the system crosses the steady-state point compared to the parameter settings that have not crossed the steady-state. This finding reveals that the system's regulatory mechanisms and temporal evolution characteristics may have undergone significant changes after bifurcation. It is not difficult to see that the construction of indices based on causal relationships is universal and applicable, not only effectively detecting temporal bifurcation phenomena but also revealing the formation of spatial patterns through spatial causal relationships, such as in the parameter area close to the occurrence of Turing bifurcation.

In summary, we tested the effectiveness of CNM on genetic networks, ecological networks, and Turing networks, respectively. They represent the tipping points of internal dynamics of a single node, the interactions between nodes, and the spatial interactions (usually caused by different levels of diffusion), accordingly. In these validations, both CNM-GC and CNM-TE showed consistently accurate and consistent results, rendering by the fact that the module of exact one eigenvalue of the discrete evolution matrix tending from $1^{-}$ to 1. However, in real datasets, their dynamics exhibit more complex characteristics, leading to various difficult to detect tipping points, such as using traditional DNB, and the identification of CNM-GC and CNM-TE may be more abundant.

\subsection{Validations on real-world datasets}

\begin{SCfigure*}
	\centering
	\includegraphics[width=0.75\textwidth]{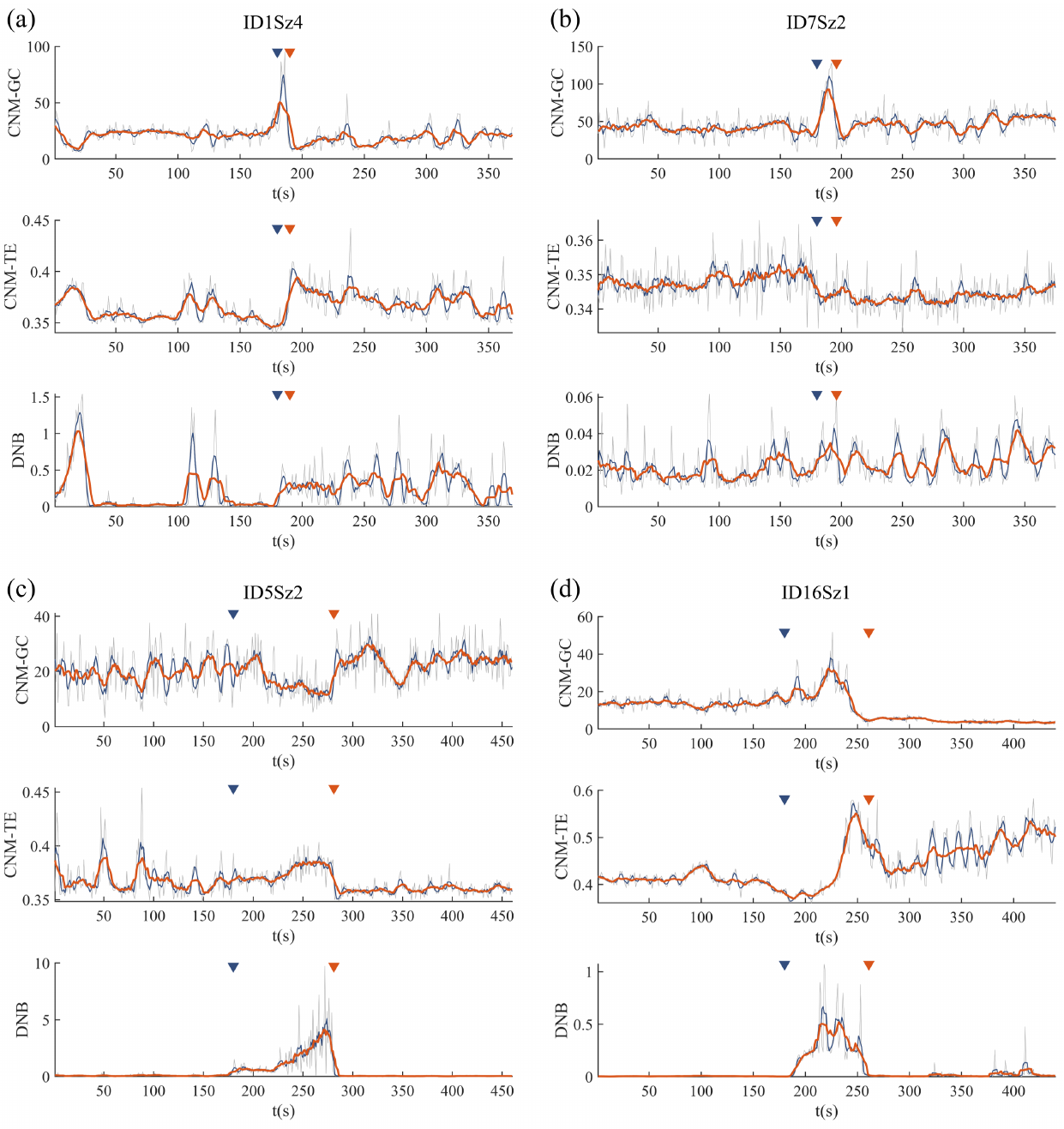}
	\caption{The tipping point detection results of iEEG short-term data set. The patients' ID and Seizure serial number in (a), (b), (c) and (d) are ID1Sz4, ID7Sz2, ID5Sz2 and ID16Sz1, accordingly. Each sub-figure consists of three subplots from top to bottom, showing the results of CNM-GC, CNM-TE and DNB, successively. The blue and red arrows in each subplot represent the disease's beginning and end, respectively. In addition, the grey line, blue line and red line in each subplot represent the original marker, the 5-second moving average of the marker, and the 12-second moving average of the marker, schematically.}
	\label{Figure_4}
\end{SCfigure*}

Although real-world datasets often have high complexity and their dynamics are difficult to construct, fortunately, mathematical theories such as the center manifold theorem ensure that the dynamics of the system near bifurcation points may be confined to a low-dimensional manifold. Therefore, network markers similar to DNB can reveal potential dynamical bifurcation phenomena in real-world datasets to a certain extent. However, the application of these methods to epilepsy data is still limited, mainly because the complex biological critical phenomena represented by epilepsy cannot be simply quantified using a few models.
In this section, we apply the two types of causal indicators we have constructed to human iEEG data from epilepsy patients, in the hope of verifying the practicality and operability of the CNM framework. Compared with the DNB-based method, our metric is more sensitive in detecting the interaction relationship between nodes and can explain the dynamic bifurcation caused by causality in both linear and nonlinear senses. Therefore, this method provides a new perspective and evaluation criteria to understand the neurodynamic characteristics of complex diseases such as epilepsy, and provides theoretical support for personalized clinical diagnosis and treatment, as the \cref{Figure_4,Table1} show.

\subsubsection{Datasets and candidate network markers introduction}

Epilepsy is a chronic disorder that affects brain function, characterized by abnormal and excessive neuronal discharges that lead to recurrent epileptic seizures. Numerous benchmark models in epilepsy research have revealed a common mechanism: epileptic seizures typically signify a critical transition in brain state \cite{Meisel2012, Negahani2015, Milanowski2016}. This transition is thought to be associated with a tipping point, where the brain rapidly shifts from a normal state to a seizure state \cite{Freestone2017, Freestone2015, Kalitzin2010}.

In the natural world, critical transitions represent pivotal moments where the behavior of a system undergoes a fundamental change, and the theory of dynamical bifurcations provides a theoretical foundation and analytical toolkit for such phenomena. This theoretical framework aids in understanding and predicting changes in behavioral patterns as a system approaches a tipping point. When applied to epilepsy research, the theory of dynamical bifurcations enables the identification and analysis of key neurodynamic changes that may precipitate epileptic seizures, offering new perspectives and methods for the prediction and intervention of seizure events.

In this part, we choose the iEEG dataset to detect the epilepsy seizures (see the detailed introduction of the datasets in Supporting Information). Meanwhile, we select DNB, CNM-GC, and CNM-TE as candidate markers for seizure prediction and employ a unified assessment standard to measure the performance of these indicators in warning of epilepsy. Since the previous indicators are incompatible with the DNB framework, we draw on the construction method of DNB indicators and apply the DG and NDG to DNB and CNMs, to assess their effectiveness on the dataset with a unified standard. Although theoretically, the corresponding DG of CNM-GC and CNM-TE is a subset of DNB, except for the causality between DG and NDG tending to 0, the impact of other pathways is a bounded constant, and these additional items will not trigger an explosion of indicators. This is crucial for our understanding of the stability and reliability of local indicators in practical application.

It is worth noting that DNB is directly influenced by the dynamics of the nodes, especially by the PCC, making it more sensitive to noise. In contrast, CNM-GC and CNM-TE are designed to capture changes in the interactions between nodes; these indicators have higher noise resistance and can keenly detect potential causality between nodes.

The division of DG and NDG on real datasets is mainly determined by the variance of the nodes, stemming from the assurance that the variance of the DG tends to infinity near the critical point. Therefore, in the epilepsy dataset, we take the average of the variance of each node during the onset period and then use the $K$-means clustering algorithm to provide group division. Note that no matter how the division is made, this division has subjectivity.

The calculation of the DNB, where
\begin{equation}
    \text{DNB(Net)} := \frac{{\rm SD}_d \cdot {\rm PCC}_d}{{\rm PCC}_o},
\end{equation}
is based on several key metrics that quantify the dynamics within and between DG and NDG. Specifically, the DNB is calculated using the following parameters:

\begin{itemize}
    \item ${\rm SD}_d$: The average standard deviation of the nodes within the DG;
    \item ${\rm PCC}_d$: The average Pearson correlation coefficient among nodes within the DG;
    \item ${\rm PCC}_o$: The average Pearson correlation coefficient between nodes in the DG and those in the NDG.
\end{itemize}

\subsubsection{Result analysis for iEEG dataset}


\begin{table*}[t!]
	\caption{The Accuracy of different combinations of markers.}\label{Table1}\centering
	\begin{tabular}{cccccc}
		\toprule[1.5pt]
		Markers & \quad All Valid \quad & \quad CNMs \quad & \quad DNB \quad & \quad CNM-GC only \quad & \quad CNM-TE only\\
		\midrule[1pt]
		Accuracy (\%) & 43        & 92          & 79          & 68       & 74\\
		\bottomrule[1.5pt]
	\end{tabular}
\end{table*}

In this section, we investigate the performance differences of various indicators on the same dataset. The results indicate that in cases where the DNB detection fails, at least one of our indicators can issue an early-warning signal during the onset of the disease. Moreover, the combination of signals from the two distinct causal network markers, CNM-GC and CNM-TE, provides additional information about epileptic seizures, that is, we can preliminarily classify epileptic seizures in terms of causality using causal indicators.

Firstly, DNB, as an efficient early-warning method, completes this warning through the variance of nodes and the PCC between nodes. However, as an early-warning tool, the effectiveness of DNB is not always reliable, especially when the nodes themselves are subject to significant noise interference. The variance term in the DNB indicator fluctuates dramatically, greatly reducing its detection power. However, causal indicators such as CNM-GC and CNM-TE are designed without considering the node's own role, thus they possess superior noise resistance. An example can be found in Supporting Information (ID7Sz2, which means the 2-nd seizure of the 7-th patient in the dataset).

Furthermore, the PCC indicator is not sensitive to minor changes in node interactions, leading to the DNB potentially being insufficient to detect early-warning signals in some cases. The CNMs use a more sensitive causal relationship for critical state early-warning, reflected by the higher accuracy of CNMs than DNB (\cref{Table1}). Specifically, the CNMs method has shown considerable effects in about 92\% of the cases, which is higher than the overall effect of DNB in about 79\%. In several seizures such as ID1Sz4 and ID7Sz2 in \cref{Figure_4}(a) and (b), the CNM indicators capture potential early-warning information more effectively than DNB. At this time, the DNB indicator shows an unstable pattern of oscillation, while the causal indicators clearly distinguish the warning peaks. Therefore, due to their sensitivity and noise resistance, CNM-GC and CNM-TE still serve as powerful tools for predicting epileptic seizures when DNB fails.

We found that the tipping dynamics of epilepsy may have different causal patterns. In terms of causality, CNM-GC characterizes the strength of linear causality, while CNM-TE describes the magnitude of nonlinear causality. The rise of CNM-GC and CNM-TE in \cref{Figure_4}(a) indicates that the epileptic seizure in this case is a mixture of linear and nonlinear causality. There are cases of epilepsy caused solely by one type of causality. As shown in \cref{Figure_4}(b), in this case, CNM-GC identifies effective early-warning, while CNM-TE fails, which indicates that the epileptic seizures in this case may be more affected by direct and linear neural activity patterns; As shown in \cref{Figure_4}(c), in this case, CNM-TE identifies effective early-warning, while CNM-GC fails. The seizures in this case may originate from more complex and nonlinear neural activity patterns.

This difference indirectly confirms the complexity of the mechanisms that trigger epileptic seizures and inspires us to conduct a more in-depth discussion and analysis of the critical dynamics of epilepsy. This may be because some high-dimensional critical dynamics properties of epilepsy cannot be characterized by low-dimensional critical dynamics, which is also worth further exploration in future work. However, in 43\% cases (\cref{Table1}), the CNM-GC, CNM-TE, and DNB indicators all show a clear upward trend, consistently demonstrating a certain sense of early-warning in the critical dynamics of epilepsy, such as \cref{Figure_4}(d). 

In addition, our causal classification align with some previous experimental validations, such as the critical slowing down (CSD) in iEEG \cite{Maturana2020}. In some rare individual cases, it reveals the existence of CSD phenomena (see Supporting Information for introduction of CSD and our identification example such as ID5Sz1). This indicates that the mode of epilepsy seizure shows more than two causal patterns we explored. It is required that we need to make a comprehensive judgment based on various indicators, to provide more accurate early-warning signals for epileptic seizures in the future.

\subsubsection{Discussion on the iEEG dataset}

This study explored the performance and potential of three markers, CNM-GC, CNM-TE, and DNB, in predicting seizures through in-depth analysis of short-term iEEG datasets. CNM and DNB effectively detect the critical dynamical patterns during seizures to a certain extent. In this section, we will summarize the advantages, problems, and prospects of CNMs in detecting epilepsy. Firstly, CNM-GC and CNM-TE have noise immunity and sensitivity. This is a significant advantage over traditional DNB methods, meaning that CNMs can maintain stable prediction performance even in data with higher system noise or unclear changes in PCC. Secondly, CNM-GC and CNM-TE bring a comprehensive evaluation. We can infer the causal patterns of critical dynamics of epileptic seizures in both linear and nonlinear senses through two different causal indicators. Furthermore, CNM is versatile. The calculation method of this indicator can be directly extended to other causal senses, such as embedded entropy (EE), which can help further comprehensively identify seizure patterns. As an attempt to identify epilepsy from a new perspective and evaluation criteria, the traditional early-warning identification methods of CNM-GC and CNM-TE may fail in some special cases due to the complexity of complex disease neurodynamics. This may be due to causal explosion in other senses besides GC and TE and may be explained by the CSD model of causal indicators. In addition, there are still many aspects to be improved in the multi-index epilepsy classifier, such as indicator selection and further experiments. At the same time, CNMs may have broad prospects for development in the clinical treatment of epilepsy. If these indicators are integrated into existing epilepsy prediction models, on the one hand, they provide causal network markers of epilepsy for researchers, and on the other hand, they may be combined with multiple indicators to provide more personalized management and treatment strategies for epilepsy patients. In general, CNM-GC and CNM-TE perform well on short-term seizure datasets with critical dynamics, and their effectiveness can be effectively verified on real-world datasets.

\section{Conclusion}

In this study, we introduce a framework of causal network markers, called CNMs, for identifying general bifurcation phenomena in the dynamical evolution of complex systems. In the framework, we construct a functional form of CNMs that reflects the strength of system causality. We introduce two markers, CNM-GC and CNM-TE, representing the ``linear causality'' and the ``non-linear causality'', respectively.

To experimentally verify the efficacy of CNM-GC and CNM-TE, we use the data produced by benchmark models and collected from the real-world systems. Precisely, we consider the data from three benchmark models: the genetic network, the ecological network, and the Turing diffusion interaction network. They respectively represent the internal, interaction, and diffusion effects reaching criticality, encompassing three types of typical critical phenomena in the common stage. The results indicate that CNMs have significant predictive effects on both temporal and spatial bifurcation models. Meanwhile, the real-world dataset we use comes from an open-source epilepsy dataset, which is typically considered to have complex dynamical properties. We found that in this situation, the identification of CNMs composed of causal indicators with different meanings may be inconsistent, and the combination of two causal network markers has high predictive accuracy, which provides a new causal perspective for early-warning and identification of epilepsy.

As a causal-oriented unidirectional markers, CNMs framework focuses on the interactions between nodes and is not affected by the internal dynamics of node variance. It can capture more information than the PCC between nodes, as in DNB, and are resistant to noise and highly sensitive. CNMs method also detects different kinds of temporal-spatial tipping points, making it simple and versatile. At the same time, it is scalable and flexible, where many of the causal framework can be introduced into this framework, including the spatial causality \cite{Gao2023NC} and the machine learning technique \cite{Zhao2024CP}. When multiple causal indicators are selected, CNMs generally provide a comprehensive evaluation, which is beneficial to identifying more complex dynamics of the system so that it is applicable to clinical disease detection and early-warning, with broad application prospects.

\medskip
\textbf{Supporting Information} \par 
Supporting Information is available from the Wiley Online Library or from the author.

\medskip
\textbf{Acknowledgements} \par 
J. S. is supported by National Key R\&D Program (Grant no. 2022YFC2704604), by the National Natural Science Foundation of China (Grant no. 12301620), and by the AI for Science Foundation of Fudan University (FudanX24AI041). W.L. is supported by the National Natural Science Foundation of China (Grant no. 11925103), by the STCSM (Grant nos. 2021SHZDZX0103, 22JC1402500, and 22JC1401402), and by the SMEC (Grant no. 2023ZKZD04). S.B. is supported by the National Natural Science Foundation of China (Grant no. 123B2021).

\medskip


\bibliographystyle{MSP}

\end{document}